\documentclass[preprint]{elsarticle}
\usepackage[utf8]{inputenc}
\usepackage{amssymb}
\usepackage{amsmath}
\usepackage{amsthm}
\usepackage{hyperref}
\usepackage{multirow}
\usepackage{a4wide}
\usepackage{breqn}
\usepackage[table,xcdraw]{xcolor}
\usepackage{xcolor}
\usepackage{mathtools}
\usepackage{bm}
\usepackage{enumitem}
\usepackage{mathrsfs}
\usepackage{graphicx}
\usepackage{caption}
\usepackage{url}
\usepackage{euscript}
\usepackage{placeins}
\usepackage{mathdots}
\usepackage{multirow}
\usepackage{arydshln} 
\usepackage[english]{babel}
\usepackage{array}
\usepackage{subcaption}
\usepackage{booktabs}
\usepackage[numbers]{natbib}
\usepackage{cases}
\usepackage{graphics}
\usepackage{graphicx}
\usepackage{algorithm}
\usepackage{algpseudocode}

\usepackage{graphics}
\usepackage[table,dvipsnames]{xcolor}
\colorlet{myColor}{RoyalBlue!20}
\usepackage[capitalize,noabbrev]{cleveref}
\usepackage{natbib}
\numberwithin{equation}{section}

\theoremstyle{remark}
\usepackage{longtable}
\usepackage{array}

\usepackage{arydshln}
\usepackage{multirow}
\usepackage{colortbl}
\usepackage{tabularray}
\usepackage{tabularx}
\usepackage[textsize=tiny]{todonotes}

\usepackage[table]{xcolor}

\begin{document}
	\begin{frontmatter}
\title{Do Existing Preconditioners Improve Biomedical Tabular Foundation Learning?\\An Empirical Study on TabPFN Optimization}
 
 	\author[inst1]{M. Sajid\corref{cor1}}
		\affiliation[inst1]{organization={Department  of Mathematics, Indian Institute of Technology Indore},
			addressline={Simrol}, 
			city={Indore},
			postcode={453552}, 
			state={Madhya Pradesh},
			country={India}}
        \cortext[cor1]{Equal contribution.}
        \cortext[cor2]{Corresponding author.}
		\ead{phd2101241003@iiti.ac.in}
		\author[inst2]{Pinki Khatun\corref{cor1}}
  \affiliation[inst2]{organization={Department of Industrial Engineering, University of Florence},
			addressline={Viale Morgagni 40/44}, 
			city={Florence},
			postcode={50134}, 
			country={Italy}}
		\ead{pinki.khatun@gunifi.it}
        		\author[inst1]{M. Tanveer\corref{cor2}}
                \ead{m.tanveer@iiti.ac.in}
	\begin{abstract}
Tabular foundation models have recently shown strong potential for structured biomedical data analysis. Among them, TabPFN has emerged as an effective approach for low-data tabular classification tasks. However, the impact of optimization and preconditioning strategies on biomedical fine-tuning remains largely unexplored. In this work, we present a comprehensive empirical investigation of five AdamW-based preconditioning strategies for fine-tuning TabPFN v2.5 on 59 biomedical datasets spanning Alzheimer’s disease, breast cancer, schizophrenia, significant memory concern (SMC), KEEL biomedical datasets, and UCI biomedical benchmarks. The evaluation considers predictive performance, computational efficiency, and statistical significance analysis. Experimental results demonstrate that the original AdamW optimizer consistently achieves the best overall performance and statistical ranking, while existing curvature-aware preconditioners fail to provide reliable improvements across diverse biomedical learning scenarios. The findings suggest that generic preconditioning approaches may not adequately capture the optimization characteristics of biomedical tabular learning, motivating the development of biomedical-aware preconditioners specifically tailored for healthcare-oriented tabular foundation models.
\end{abstract}

\end{frontmatter}

\section{Introduction}

The rapid emergence of tabular foundation models has fundamentally reshaped modern machine learning, enabling pretrained architectures to generalize across diverse downstream tasks with minimal task-specific adaptation \cite{hollmann2025accurate, yang2024unitabe}. In the domain of tabular learning, TabPFN \cite{hollmann2025accurate, hollmanntabpfn} has recently introduced a paradigm shift by reformulating tabular prediction through prior-data fitted transformers trained over a large distribution of synthetic tasks. Unlike conventional tabular models that depend heavily on dataset-specific optimization and extensive hyperparameter tuning \cite{gorishniy2021revisiting, arik2021tabnet, grinsztajn2022tree}, TabPFN leverages pretrained probabilistic priors to achieve remarkable predictive performance, particularly in low-data regimes. Owing to its strong generalization ability and effectiveness in low-data regimes, TabPFN has recently attracted increasing attention \cite{qu2025tabicl, buhlertowards, robertsonfairpfn} and has found its application in biomedical machine learning \cite{liang2025tabular}, including Alzheimer's disease diagnosis \cite{ye2026evaluating}, drug discovery \cite{chen2026tabpfn}, and RNA-seq analysis \cite{zhou2025limitations}.

Despite these advances, a critical and largely unexplored challenge remains unresolved: how should pretrained tabular foundation models be optimized during downstream biomedical adaptation? Existing research predominantly focuses on architectural improvements, feature engineering, or data-centric preprocessing strategies, while optimization itself is often treated as a secondary implementation detail. However, this assumption becomes highly problematic in biomedical learning scenarios, where datasets are typically characterized by small sample sizes, high-dimensional biomarkers, noisy measurements, class imbalance, and substantial inter-subject variability \cite{liang2025tabular}. Under such challenging conditions, the optimization strategy directly influences convergence stability, representation preservation, calibration reliability, and the generalization capability of the model \cite{thomas2019deep, sun2019optimization}.

This challenge becomes even more important for transformer-based tabular foundation models due to their highly non-convex optimization landscapes and complex parameter interactions \cite{lourencco2026context}. While standard adaptive optimizers such as AdamW \cite{loshchilov2017decoupled} are widely adopted in practice \cite{loshchilov2017decoupled}, they primarily rely on diagonal gradient statistics and may fail to capture richer curvature structures embedded within pretrained representations \cite{gupta2018shampoo,martens2020new,anil2020scalable}. Consequently, the extent to which different curvature-aware preconditioning strategies affect the biomedical adaptation of TabPFN remains an open and fundamentally important research question.

Motivated by these challenges, we conduct a systematic investigation of AdamW-based preconditioning strategies for fine-tuning TabPFN on biomedical classification tasks. Rather than treating optimization solely as a training utility, we study preconditioning as a key factor governing the adaptation behavior, stability, and generalization of pretrained tabular foundation models in biomedical settings. To this end, five representative AdamW-based optimization variants are analyzed, including standard AdamW \cite{loshchilov2017decoupled}, diagonal curvature approximation \cite{duchi2011adaptive,kingma2014adam}, Fisher-guided preconditioning \cite{martens2020new}, quasi-Newton approximation \cite{liu1989limited}, and structured matrix-aware optimization \cite{gupta2018shampoo}. Experiments are performed across 59 biomedical datasets spanning diverse healthcare domains, feature dimensions, sample sizes, and imbalance characteristics. 

\noindent \textbf{The main contributions of this work are summarized as:}
\begin{itemize}[leftmargin=*]
    \item We present a comprehensive empirical study of AdamW-based preconditioning strategies for fine-tuning the \textit{TabPFN v2.5} \cite{hollmann2025accurate, tabpfn_repo} foundation model on biomedical tasks.
    \item We establish a unified optimization benchmark involving five representative AdamW-based variants, including diagonal, Fisher-guided, quasi-Newton, and structured matrix-aware preconditioning methods.
    \item We conduct extensive evaluations on 59 biomedical datasets spanning Alzheimer’s disease (AD), breast cancer, schizophrenia, significant memory concern (SMC), KEEL biomedical datasets, and UCI biomedical benchmarks with diverse characteristics and complexities.
    \item Experimental and statistical analyses demonstrate that existing generic preconditioning approaches do not consistently surpass the original AdamW optimizer for biomedical tabular foundation learning, emphasizing the necessity of designing domain-specific biomedical preconditioners for reliable healthcare AI systems.
\end{itemize}

The remainder of this paper is organized as follows. Section~\ref{Sec:optimizer_details} reviews the related work and relevant literature. Section~\ref{Sec:Benchmark_Approach} describes the experimental configuration, benchmark setup, dataset details, preprocessing procedures, and evaluation metrics. Section~\ref{Sec:Results_and_Discussion} presents the experimental results, statistical analysis, and practical recommendations. Finally, Section~\ref{Sec:Conclusion} concludes the paper.

\section{Related Works and Literature}
\label{Sec:optimizer_details}
This section provides an overview of the related literature and the mathematical background of the baseline preconditioners.

\subsection{Tabular Foundation Models and Curvature-Aware Optimization}

Foundation models have recently transformed tabular machine learning by enabling pretrained architectures to generalize across diverse downstream tasks. Among these approaches, \textit{TabPFN} \cite{hollmann2025accurate, hollmanntabpfn} reformulates tabular prediction through prior-data fitted transformers trained over a large distribution of synthetic tasks. Given training data $(X_{\text{train}},y_{\text{train}})$ and test samples $X_{\text{test}}$, TabPFN estimates
\[
p(y_{\text{test}} \mid X_{\text{train}}, y_{\text{train}}, X_{\text{test}}),
\]
thereby approximating Bayesian inference through transformer-based pretraining.

Unlike conventional tabular learning models that require extensive hyperparameter optimization, TabPFN leverages pretrained probabilistic priors for efficient low-data prediction. Owing to its strong generalization capability, TabPFN has recently attracted increasing attention in biomedical applications such as disease diagnosis, biomarker analysis, and clinical prediction. However, most existing studies primarily emphasize predictive performance, while the optimization behavior of TabPFN during downstream fine-tuning remains largely unexplored.

Transformer-based tabular foundation models exhibit highly non-convex optimization landscapes with strong parameter interactions \cite{lourencco2026context}. Consequently, optimization strategies can substantially influence convergence stability, calibration reliability, and downstream generalization, particularly in biomedical settings characterized by high-dimensional features, noisy measurements, limited sample sizes, and severe class imbalance \cite{thomas2019deep, sun2019optimization}.

Optimization and preconditioning strategies therefore play a critical role in deep neural network training. AdamW is one of the most widely adopted adaptive optimizers for transformer models, updating parameters as
\[
\theta_{t+1} = \theta_t - \eta \frac{\hat{m}_t}{\sqrt{\hat{v}_t}+\epsilon} - \eta \lambda \theta_t,
\]
where $\hat{m}_t$ and $\hat{v}_t$ denote the bias-corrected first and second moment estimates. Although effective, AdamW mainly relies on diagonal gradient statistics and may fail to capture richer curvature structures.

To address this limitation, several curvature-aware preconditioning methods have been proposed. Diagonal preconditioners estimate parameter-wise curvature using local second-order statistics \cite{duchi2011adaptive,kingma2014adam}, whereas quasi-Newton methods such as L-BFGS construct low-rank approximations of the inverse Hessian from historical gradient updates \cite{liu1989limited}. Fisher-guided approaches approximate the Fisher Information Matrix \cite{martens2020new}
\[
F = \mathbb{E}\left[\nabla_\theta \log p(y|x;\theta)\nabla_\theta \log p(y|x;\theta)^T\right],
\]
to better characterize optimization geometry. Structured matrix-aware methods such as Shampoo further exploit parameter correlations through matrix preconditioning \cite{gupta2018shampoo}:
\[
W_{t+1} = W_t - \eta L_t^{-\frac{1}{4}} G_t R_t^{-\frac{1}{4}},
\]
where $G_t$ denotes the gradient matrix and $L_t$, $R_t$ represent layer-wise covariance statistics.

Despite substantial advances in curvature-aware optimization, their effectiveness for pretrained tabular foundation models remains insufficiently studied, particularly in biomedical fine-tuning scenarios. Motivated by this gap, the present study systematically investigates five AdamW-based optimization variants for TabPFN fine-tuning: a) AdamW \cite{loshchilov2017decoupled}, b) AdamW-Diag \cite{duchi2011adaptive,kingma2014adam}, c) AdamW-Fisher \cite{martens2020new}, d) AdamW-LBFGS \cite{liu1989limited}, and e) AdamW-Shampoo \cite{gupta2018shampoo}.

These optimizers collectively span multiple levels of curvature approximation, ranging from lightweight adaptive scaling to structured second-order geometric modeling. The selected preconditioners possess strong theoretical foundations, practical scalability for transformer optimization, and complementary mechanisms for capturing diagonal, low-rank, Fisher-guided, and matrix-aware curvature information.

\subsection{Mathematical Formulations of the Optimizers}
This section provides an overview of the related literature and the mathematical background of the preconditioners.\\

\noindent \textbf{\textit{AdamW Baseline.}} AdamW combines adaptive moment estimation with decoupled weight decay regularization:
\begin{equation}
m_t = \beta_1 m_{t-1} + (1-\beta_1)g_t,
\end{equation}

\begin{equation}
v_t = \beta_2 v_{t-1} + (1-\beta_2)g_t^2,
\end{equation}

\begin{equation}
\theta_{t+1} = \theta_t - \alpha \frac{\hat{m}_t}{\sqrt{\hat{v}_t}+\epsilon} - \lambda \theta_t.
\end{equation}

\noindent \textbf{\textit{AdamW-Diag.}} AdamW-Diag employs a diagonal Hessian approximation using second-order gradient statistics:
\begin{equation}
D_t = \text{diag}(v_t + \epsilon),
\end{equation}

\begin{equation}
g_{\text{precond}} = \frac{g_t}{\sqrt{D_t}}.
\end{equation}

This formulation introduces element-wise adaptive scaling with low computational complexity.\\

\noindent \textbf{\textit{AdamW-Fisher.}} AdamW-Fisher utilizes a diagonal Fisher Information Matrix approximation:
\begin{equation}
F_t = \gamma F_{t-1} + (1-\gamma)g_t^2,
\end{equation}

\begin{equation}
g_{\text{precond}} = \frac{g_t}{\sqrt{F_t+\epsilon}},
\end{equation}
where $\gamma$ denotes the Fisher decay coefficient.\\

\noindent \textbf{\textit{AdamW-LBFGS.}} AdamW-LBFGS incorporates limited-memory quasi-Newton updates through curvature pairs $(s_k,y_k)$:
\begin{equation}
s_k = \theta_k - \theta_{k-1},
\end{equation}

\begin{equation}
y_k = \nabla f_k - \nabla f_{k-1}.
\end{equation}

The inverse Hessian approximation is computed efficiently using the standard two-loop recursion strategy.\\

\noindent \textbf{\textit{AdamW-Shampoo.}} AdamW-Shampoo leverages Kronecker-factored curvature approximations:
\begin{equation}
G_{\text{precond}} = L_t^{-1/4} G_t R_t^{-1/4},
\end{equation}
where $L_t$ and $R_t$ denote the left and right second-order statistics of the gradient matrix.

\section{Benchmark Approach}\label{Sec:Benchmark_Approach}


In this study, we use the TabPFN v2.5 \cite{hollmann2025accurate, tabpfn_repo} framework as the backbone architecture and evaluates five AdamW-based preconditioner variants, including AdamW \cite{loshchilov2017decoupled}, AdamW-Diag based on diagonal curvature approximation \cite{duchi2011adaptive,kingma2014adam}, AdamW-Fisher using Fisher-guided preconditioning \cite{martens2020new}, AdamW-LBFGS employing quasi-Newton approximation \cite{liu1989limited}, and AdamW-Shampoo utilizing structured matrix-aware optimization \cite{gupta2018shampoo}. The study investigates the effectiveness of these preconditioning strategies during fine-tuning across diverse biomedical classification datasets.  


\subsection{Experimental Configuration and Setup}\label{Sec:Appendix_Configuration}

All experiments are implemented in Python 3.12.3 using the PyTorch 2.5.1 framework with CUDA 12.1 support. In this study, we use the TabPFN v2.5 framework as the backbone architecture and evaluate multiple AdamW-based preconditioner variants. The proposed models and all AdamW-based preconditioner variants are trained on a workstation running Microsoft Windows 11 Pro Education. The experimental platform consists of dual Intel Xeon processors operating at approximately 2.90 GHz with 256 GB RAM. Model training and inference are performed using a single NVIDIA RTX A4500 GPU with 20 GB dedicated memory. 

The implementation utilizes GPU-accelerated computation through CUDA-enabled PyTorch operations to ensure efficient large-scale experimentation across all biomedical datasets. The software environment additionally includes standard scientific computing libraries such as NumPy, SciPy, Pandas, and Scikit-learn for data processing, evaluation, and statistical analysis.

GPU memory utilization is monitored using \texttt{torch.cuda.max\_memory\_allocated()}, whereas RAM consumption is tracked using \texttt{tracemalloc}. These efficiency metrics facilitate a comprehensive analysis of the trade-offs between predictive performance and computational overhead across different curvature-aware optimization strategies.

All experiments are conducted using five independent runs with random seeds $\{41,42,43,44,45\}$ to ensure statistical reliability and reproducibility. For each dataset, an 80:20 stratified train-test split is employed, followed by feature normalization using the StandardScaler transformation. The final performance is reported as mean$\pm$standard deviation across all runs.

The training configuration remains fixed across all optimizer variants to ensure fair comparison. Table~\ref{tab:hyperparameter_settings} summarizes the hyperparameter settings used throughout the experiments and the best configuration related to preconditioner employed in the study is given in Table \ref{tab:preconditioner_config}.

\begin{table}[!t]
\centering
\caption{Hyperparameter settings used for fine-tuning TabPFN across all optimizer variants.}
\label{tab:hyperparameter_settings}
\resizebox{7cm}{!}{
\begin{tabular}{lc}
\toprule
\textbf{Hyperparameter} & \textbf{Value} \\
\midrule
Epochs & 20 \\
Learning Rate & $5\times10^{-4}$ \\
Weight Decay & 0.01 \\
Validation Split & 0.1 \\
Early Stopping Patience & 3 \\
Gradient Clipping & 1.0 \\
Learning Rate Scheduler & Cosine Annealing \\
Context-Query Samples & 1000 \\
Context-Query Split Ratio & 0.2 \\
Random Seeds & $\{41,42,43,44,45\}$ \\
\bottomrule
\end{tabular}
}
\end{table}

\begin{table}[!htbp]
\centering
\caption{Best configurations related to preconditioner employed in the study.}
\label{tab:preconditioner_config}
\begin{tabular}{ll}
\toprule
\textbf{Optimizer} & \textbf{Configuration} \\
\midrule
AdamW & No preconditioner \\
AdamW-Diag & $\epsilon=10^{-8}, \beta_2=0.999$ \\
AdamW-Fisher & $\epsilon=10^{-8}, \beta_2=0.999,$ Fisher decay $=0.95$ \\
AdamW-LBFGS & Memory size $=50,$ $\epsilon=10^{-8}$ \\
AdamW-Shampoo & $\epsilon=10^{-8}, \beta_2=0.9$ \\
\bottomrule
\end{tabular}
\end{table}

\subsection{Dataset Details and Preprocessing}

The benchmark suite consists of 59 datasets grouped into seven major biomedical categories: Alzheimer’s Disease (3 datasets)\footnote{\url{https://adni.loni.usc.edu}}, Breast Cancer (16 datasets) \cite{spanhol2015dataset}, KEEL Biomedical (6 datasets) \cite{derrac2015keel}, Schizophrenia (3 datasets)\footnote{\url{http://fcon_1000.projects.nitrc.org/indi/retro/cobre.html}}, Significant Memory Concern (SMC) (5 datasets)\footnote{\url{https://adni.loni.usc.edu}} \cite{sajid2024decoding}, UCI Binary Biomedical (16 datasets), and UCI Multiclass Biomedical (10 datasets) \cite{dua2017uci}. These datasets cover diverse biomedical domains, including oncology, cardiology, neurology, hepatology, endocrinology, dermatology, radiology, hematology, fetal cardiology, orthopedic disorders, and schizophrenia analysis. The datasets exhibit substantial variability in terms of sample size, feature dimensionality, and class imbalance, enabling a comprehensive evaluation under heterogeneous biomedical learning conditions. The benchmark includes both small-scale and large-scale datasets, with sample sizes ranging from 32 to 7200 and feature dimensions varying from 3 to 889. In addition, the datasets include both balanced and highly imbalanced scenarios, with imbalance ratios ranging from 1.0 to over 40. See Table \ref{tab:dataset_summary} for statistical details of the datasets, and detailed preprocessing is provided below.

\subsubsection{Alzheimer’s Disease Neuroimaging Initiative (ADNI) Dataset} Alzheimer’s disease (AD) is a progressive neurodegenerative disorder that gradually deteriorates memory, cognitive functioning, and behavioral abilities. To evaluate the proposed framework on neuroimaging-based Alzheimer’s disease diagnosis, structural MRI scans from the Alzheimer’s Disease Neuroimaging Initiative (ADNI) database\footnote{\url{https://adni.loni.usc.edu}} are utilized. The ADNI project was launched in 2003 as a large-scale longitudinal initiative aimed at investigating biomarkers and neuroimaging techniques, including magnetic resonance imaging (MRI), positron emission tomography (PET), genetic assessments, and clinical evaluations, for the early detection and progression analysis of Alzheimer’s disease and mild cognitive impairment (MCI). In this study, three binary classification settings are considered, namely cognitively normal (CN) versus Alzheimer’s disease (AD), cognitively normal (CN) versus mild cognitive impairment (MCI), and mild cognitive impairment (MCI) versus Alzheimer’s disease (AD). The CN-vs-AD dataset contains 415 subjects with 228 cognitively normal individuals and 187 AD patients. The CN-vs-MCI dataset consists of 626 subjects, including 398 CN subjects and 228 MCI subjects. The MCI-vs-AD dataset includes 585 subjects comprising 398 MCI subjects and 187 AD patients. Each dataset contains 91 extracted structural features. The preprocessing and feature extraction pipeline follows the methodology described in \cite{richhariya2020diagnosis, ganaie2024graph}. Structural MRI scans are processed using standard neuroimaging preprocessing procedures, including bias correction, tissue segmentation, spatial normalization, and smoothing. Subsequently, region-of-interest (ROI)-based structural features are extracted from anatomically relevant brain regions for downstream classification experiments.

\subsubsection{Breast Cancer Histopathology Dataset} 
The breast cancer histopathology dataset consists of 1,240 histopathological image scans acquired at 400$\times$ magnification and categorized into benign and malignant breast tissue classes \cite{spanhol2015dataset}. The benign subclasses include adenosis (106 samples), fibroadenoma (237 samples), phyllodes tumor (115 samples), and tubular adenoma (130 samples). The malignant subclasses comprise lobular carcinoma (137 samples), papillary carcinoma (138 samples), ductal carcinoma (208 samples), and mucinous carcinoma (169 samples). Feature extraction is performed following the methodology described in \cite{gautam2020minimum}.

\subsubsection{KEEL Biomedical Datasets} The KEEL biomedical benchmark\footnote{\url{https://sci2s.ugr.es/keel/datasets.php}} \cite{derrac2015keel} suite includes six widely used medical datasets spanning multiple healthcare domains, including oncology, cardiology, endocrinology, hematology, and survival analysis. Specifically, the datasets comprise Wisconsin Breast Cancer, Cleveland Heart Disease, Haberman Survival, Pima Diabetes, and Blood Transfusion datasets. The datasets vary considerably in terms of sample size, feature dimensionality, and imbalance characteristics, with the number of samples ranging from 297 to 768 and feature dimensions varying from 3 to 13. These datasets provide diverse clinical learning scenarios for evaluating the robustness and generalization capability of the proposed framework under heterogeneous biomedical conditions.

\begin{table*}[htbp]
\centering
\caption{Summary of the datasets used for comprehensive evaluation, including abbreviation, subdomain, total samples, feature dimensions, majority/minority class distributions, and imbalance ratio (IR).}
\label{tab:dataset_summary}
\resizebox{15cm}{!}{
\begin{tabular}{llllllll}
\toprule
\textbf{Dataset} & \textbf{Abbreviation} & \textbf{Subdomain} & \textbf{\#Total Samples} & \textbf{\#Features} & \textbf{\#Majority} & \textbf{\#Minority} & \textbf{IR} \\
\midrule

\multicolumn{8}{c}{\textbf{Alzheimer's Disease (AD) $|$ No. of Datasets: 03}} \\
\midrule
CN\_vs\_AD & CN\_vs\_AD & -- & 415 & 91 & 228 & 187 & 1.2193 \\
CN\_vs\_MCI & CN\_vs\_MCI & -- & 626 & 91 & 398 & 228 & 1.7456 \\
MCI\_vs\_AD & MCI\_vs\_AD & -- & 585 & 91 & 398 & 187 & 2.1283 \\
\midrule

\multicolumn{8}{c}{\textbf{Breast Cancer $|$ No. of Datasets: 16}} \\
\midrule
adenosis\_vs\_ductal\_carcinoma & ADC & -- & 314 & 768 & 208 & 106 & 1.9623 \\
adenosis\_vs\_lobular\_carcinoma & ALC & -- & 243 & 768 & 137 & 106 & 1.2925 \\
adenosis\_vs\_mucinous\_carcinoma & AMC & -- & 275 & 768 & 169 & 106 & 1.5943 \\
adenosis\_vs\_papillary\_carcinoma & APC & -- & 244 & 768 & 138 & 106 & 1.3019 \\
fibroadenoma\_vs\_ductal\_carcinoma & FDC & -- & 445 & 768 & 237 & 208 & 1.1394 \\
fibroadenoma\_vs\_lobular\_carcinoma & FLC & -- & 374 & 768 & 237 & 137 & 1.7299 \\
fibroadenoma\_vs\_mucinous\_carcinoma & FMC & -- & 406 & 768 & 237 & 169 & 1.4024 \\
fibroadenoma\_vs\_papillary\_carcinoma & FPC & -- & 375 & 768 & 237 & 138 & 1.7174 \\
phyllodes\_tumour\_vs\_ductal\_carcinoma & PTDC & -- & 323 & 768 & 208 & 115 & 1.8087 \\
phyllodes\_tumour\_vs\_lobular\_carcinoma & PTLC & -- & 252 & 768 & 137 & 115 & 1.1913 \\
phyllodes\_tumour\_vs\_mucinous\_carcinoma & PTMC & -- & 284 & 768 & 169 & 115 & 1.4696 \\
phyllodes\_tumour\_vs\_papillary\_carcinoma & PTPC & -- & 253 & 768 & 138 & 115 & 1.2000 \\
tubular\_adenoma\_vs\_ductal\_carcinoma & TADC & -- & 338 & 768 & 208 & 130 & 1.6000 \\
tubular\_adenoma\_vs\_lobular\_carcinoma & TALC & -- & 267 & 768 & 137 & 130 & 1.0538 \\
tubular\_adenoma\_vs\_mucinous\_carcinoma & TAMC & -- & 299 & 768 & 169 & 130 & 1.3000 \\
tubular\_adenoma\_vs\_papillary\_carcinoma & TAPC & -- & 268 & 768 & 138 & 130 & 1.0615 \\
\midrule

\multicolumn{8}{c}{\textbf{KEEL Biomedical $|$ No. of Datasets: 06}} \\
\midrule
brwisconsin & BRW & Oncology & 683 & 9 & 444 & 239 & 1.8577 \\
cleve & CLV & Cardiology & 297 & 13 & 160 & 137 & 1.1679 \\
haber & HAB & Survival Analysis (Breast Cancer) & 306 & 3 & 225 & 81 & 2.7778 \\
haberman & HBMAN & Survival Analysis (Breast Cancer) & 306 & 3 & 225 & 81 & 2.7778 \\
pima & PIMA & Endocrinology (Diabetes) & 768 & 8 & 500 & 268 & 1.8657 \\
transfusion & TRAN & Hematology (Blood Transfusion) & 748 & 4 & 570 & 178 & 3.2022 \\
\midrule

\multicolumn{8}{c}{\textbf{Schizophrenia $|$ No. of Datasets: 03}} \\
\midrule
Schozophrenia\_ROI\_Combined\_GM\_WM & SRCGW & -- & 146 & 272 & 74 & 72 & 1.0278 \\
Schozophrenia\_ROI\_GM & SRG & -- & 146 & 136 & 74 & 72 & 1.0278 \\
Schozophrenia\_ROI\_WM & SRW & -- & 146 & 136 & 74 & 72 & 1.0278 \\
\midrule

\multicolumn{8}{c}{\textbf{Significant Memory Concern (SMC) $|$ No. of Datasets: 05}} \\
\midrule
All\_feature\_combined & ALL FEAT & -- & 222 & 889 & 111 & 111 & 1.0000 \\
CT\_feature & CT & -- & 222 & 70 & 111 & 111 & 1.0000 \\
GM\_feature & GM & -- & 222 & 275 & 111 & 111 & 1.0000 \\
WJ\_feature & WJ & -- & 222 & 275 & 111 & 111 & 1.0000 \\
WM\_feature & WM & -- & 222 & 275 & 111 & 111 & 1.0000 \\
\midrule

\multicolumn{8}{c}{\textbf{UCI Binary Biomedical $|$ No. of Datasets: 16}} \\
\midrule
acute\_inflammation & ACI & Urology / Inflammation & 120 & 6 & 61 & 59 & 1.0339 \\
acute\_nephritis & ACN & Nephrology & 120 & 6 & 70 & 50 & 1.4000 \\
breast\_cancer & BRC & Oncology & 286 & 9 & 201 & 85 & 2.3647 \\
breast\_cancer\_wisc & BRCW & Oncology & 699 & 9 & 458 & 241 & 1.9004 \\
breast\_cancer\_wisc\_diag & BRCWD & Oncology (Diagnostic) & 569 & 30 & 357 & 212 & 1.6840 \\
breast\_cancer\_wisc\_prog & BRCWP & Oncology (Prognostic) & 198 & 33 & 151 & 47 & 3.2128 \\
echocardiogram & ECHO & Cardiology (Imaging) & 131 & 10 & 88 & 43 & 2.0465 \\
haberman\_survival & HAS & Survival Analysis (Breast Cancer) & 306 & 3 & 225 & 81 & 2.7778 \\
heart\_hungarian & HEART & Cardiology & 294 & 12 & 188 & 106 & 1.7736 \\
hepatitis & HEPAT & Hepatology & 155 & 19 & 123 & 32 & 3.8438 \\
horse\_colic & HOC & Veterinary Medicine / Gastroenterology & 368 & 25 & 232 & 136 & 1.7059 \\
ilpd\_indian\_liver & ILIL & Hepatology & 583 & 9 & 416 & 167 & 2.4910 \\
mammographic & MAMMO & Radiology (Breast Imaging) & 961 & 5 & 516 & 445 & 1.1596 \\
parkinsons & PARKIN & Neurology (Parkinson’s Disease) & 195 & 22 & 147 & 48 & 3.0625 \\
pima & PIMA & Endocrinology (Diabetes) & 768 & 8 & 500 & 268 & 1.8657 \\
vertebral\_column\_2clases & VEC2C & Orthopedics / Spine Disorders & 310 & 6 & 210 & 100 & 2.1000 \\
\midrule

\multicolumn{8}{c}{\textbf{UCI Multiclass Biomedical $|$ No. of Datasets: 10}} \\
\midrule
breast\_tissue & BRT & Oncology (Breast Tissue Imaging) & 106 & 9 & 22 & 14 & 1.5714 \\
cardiotocography\_10clases & CA10C & Obstetrics / Fetal Cardiology & 2126 & 21 & 579 & 53 & 10.9245 \\
cardiotocography\_3clases & CA3C & Obstetrics / Fetal Cardiology & 2126 & 21 & 1655 & 176 & 9.4034 \\
dermatology & DER & Dermatology & 366 & 34 & 112 & 20 & 5.6000 \\
heart\_cleveland & HEC & Cardiology & 303 & 13 & 164 & 13 & 12.6154 \\
heart\_switzerland & HES & Cardiology & 123 & 12 & 48 & 5 & 9.6000 \\
heart\_va & HEV & Cardiology & 200 & 12 & 56 & 10 & 5.6000 \\
lung\_cancer & LUC & Oncology (Lung Cancer) & 32 & 56 & 13 & 9 & 1.4444 \\
thyroid & THY & Endocrinology (Thyroid Disorders) & 7200 & 21 & 6666 & 166 & 40.1566 \\
vertebral\_column\_3clases & VEC3C & Orthopedics / Spine Disorders & 310 & 6 & 150 & 60 & 2.5000 \\
\bottomrule
\end{tabular}
}
\end{table*}

\subsubsection{Schizophrenia ROI Dataset}
The schizophrenia ROI dataset is obtained from the COBRE repository\footnote{\url{http://fcon_1000.projects.nitrc.org/indi/retro/cobre.html}} and consists of 72 schizophrenia subjects (age range: 18--65 years; mean age $38.1 \pm 13.9$ years) and 74 healthy controls (mean age $35.8 \pm 11.5$ years). ROI-based structural features are extracted following the preprocessing and feature extraction pipeline described in \cite{tanveer2022intuitionistic}.

\subsubsection{Significant Memory Concern (SMC)}\label{sec:smc}

\noindent Structural magnetic resonance imaging (MRI) data obtained from the Alzheimer’s Disease Neuroimaging Initiative (ADNI2)\footnote{\url{https://adni.loni.usc.edu}} are utilized to investigate significant memory concern (SMC) (also known as subjective cognitive decline (SCD)). We preprocess data using the process mentioned in \cite{sajid2024decoding}. The cohort consists of 111 individuals diagnosed with SCD (mean age: $72.31 \pm 5.49$ years; 57\% female) and 111 cognitively healthy control (HC) subjects (mean age: $73.36 \pm 6.36$ years; 53\% female). Statistical analysis indicates no significant differences between the groups in terms of age and sex distribution. T1-weighted MRI scans are preprocessed using the Computational Anatomy Toolbox (CAT12) integrated within Statistical Parametric Mapping (SPM12). The preprocessing framework begins with bias field correction to minimize intensity non-uniformities, followed by tissue segmentation into gray matter (GM), white matter (WM), and cerebrospinal fluid (CSF) components. Subsequently, high-dimensional Diffeomorphic Anatomical Registration Through Exponentiated Lie Algebra (DARTEL) normalization is employed to align all subjects to the Montreal Neurological Institute (MNI) standard anatomical space using an isotropic voxel resolution of 1.5 mm$^3$. Modulation is incorporated during spatial normalization to preserve local volumetric characteristics. Jacobian determinant (JD) maps are additionally generated to quantify regional tissue deformation patterns corresponding to local expansion and contraction. To improve spatial consistency and reduce noise artifacts, Gaussian smoothing with a full width at half maximum (FWHM) of 4 mm is applied to GM, WM, and JD maps. Image quality and segmentation reliability are further verified through visual inspection and CAT12 homogeneity assessment procedures. Regional structural descriptors are extracted from GM, WM, and JD representations using the Brainnetome atlas, resulting in 273 quantitative imaging features for each subject. Cortical thickness (CT) measurements are computed using the Desikan–Killiany–Tourville (DKT) atlas, producing 68 cortical features. In addition, demographic attributes including age and sex are incorporated into each modality-specific feature representation prior to model development and evaluation.

\subsubsection{UCI Binary Biomedical Datasets} The UCI binary biomedical benchmark \cite{dua2017uci} suite consists of 16 datasets collected from diverse biomedical and healthcare domains, including oncology, cardiology, hepatology, endocrinology, neurology, nephrology, radiology, orthopedics, and survival analysis. The datasets exhibit substantial variability in terms of sample size, feature dimensionality, and class imbalance characteristics, with the number of samples ranging from 120 to 961 and the number of features varying between 3 and 33. Furthermore, the imbalance ratio spans from nearly balanced datasets such as Acute Inflammation (IR = 1.03) to highly imbalanced datasets such as Hepatitis and Parkinson’s disease datasets with imbalance ratios exceeding 3.0. This diversity enables a comprehensive assessment of the proposed framework under heterogeneous binary biomedical classification scenarios.

\subsubsection{UCI Multiclass Biomedical Datasets} The UCI multiclass biomedical benchmark \cite{dua2017uci} suite contains 10 datasets covering multiple clinical domains, including cardiology, endocrinology, dermatology, oncology, fetal cardiology, and orthopedic disorders. These datasets present more challenging multiclass learning settings with varying levels of feature dimensionality and severe class imbalance. The datasets range from small-scale problems such as Lung Cancer with only 32 samples to large-scale datasets such as Thyroid containing 7200 samples, while the feature dimensions vary from 6 to 56. Additionally, the imbalance ratio ranges from 1.44 to 40.15, highlighting the presence of highly skewed multiclass distributions. Such diverse multiclass biomedical datasets provide a rigorous evaluation environment for analyzing the effectiveness and robustness of the proposed optimization framework across complex real-world clinical classification tasks.

\begin{table*}[!t]
\centering
\caption{Average performance and computational efficiency comparison of different AdamW-based preconditioners across biomedical datasets. Results are reported as mean$\pm$std.}
\label{tab:overall_avg_results}
\resizebox{\textwidth}{!}{
\begin{tabular}{lcccccc}
\toprule
\textbf{Category} &
\textbf{Metric} &
\textbf{AdamW} &
\textbf{AdamW-Diag} &
\textbf{AdamW-Fisher} &
\textbf{AdamW-LBFGS} &
\textbf{AdamW-Shampoo} \\
\midrule

\textbf{Alzheimer's Disease} &
\begin{tabular}[c]{@{}c@{}}
Accuracy \\
Sensitivity \\
Precision \\
TrainTime (s) \\
InferenceTime (s) \\
PeakGPU (MB)
\end{tabular}
&
\begin{tabular}[c]{@{}c@{}}
\cellcolor{myColor}\textbf7{6.52$\pm$9.21 }\\
\cellcolor{myColor}\textbf{68.38$\pm$24.03} \\
73.22$\pm$13.86 \\
10.68$\pm$4.25 \\
0.86$\pm$0.09 \\
\cellcolor{myColor}\textbf{792.21$\pm$93.49}
\end{tabular}
&
\begin{tabular}[c]{@{}c@{}}
74.37$\pm$9.23 \\
66.21$\pm$24.70 \\
71.71$\pm$16.40 \\
10.06$\pm$2.56 \\
0.84$\pm$0.04 \\
833.10$\pm$93.49
\end{tabular}
&
\begin{tabular}[c]{@{}c@{}}
74.35$\pm$9.34 \\
66.37$\pm$24.81 \\
71.67$\pm$16.49 \\
10.09$\pm$2.40 \\
0.84$\pm$0.06 \\
833.11$\pm$93.50
\end{tabular}
&
\begin{tabular}[c]{@{}c@{}}
74.52$\pm$9.46 \\
67.84$\pm$25.23 \\
71.05$\pm$16.33 \\
\cellcolor{myColor}\textbf{9.83$\pm$2.20} \\
0.85$\pm$0.08 \\
983.02$\pm$103.83
\end{tabular}
&
\begin{tabular}[c]{@{}c@{}}
75.42$\pm$9.62 \\
68.32$\pm$22.49 \\
\cellcolor{myColor}\textbf{73.36$\pm$16.02} \\
10.09$\pm$2.04 \\
\cellcolor{myColor}\textbf{0.83$\pm$0.03 }\\
827.68$\pm$93.50
\end{tabular}
\\
\midrule

\textbf{Breast Cancer} &
\begin{tabular}[c]{@{}c@{}}
Accuracy \\
Sensitivity \\
Precision \\
TrainTime (s) \\
InferenceTime (s) \\
PeakGPU (MB)
\end{tabular}
&
\begin{tabular}[c]{@{}c@{}}
\cellcolor{myColor}\bf 66.02$\pm$11.36 \\
51.32$\pm$36.97 \\
\cellcolor{myColor}\bf 54.86$\pm$30.47 \\
28.10$\pm$8.44 \\
1.66$\pm$0.22 \\
\cellcolor{myColor}\bf 2154.88$\pm$294.25
\end{tabular}
&
\begin{tabular}[c]{@{}c@{}}
59.55$\pm$10.02 \\
52.85$\pm$20.32 \\
53.08$\pm$16.58 \\
\cellcolor{myColor}\bf 19.70$\pm$3.91 \\
\cellcolor{myColor}\bf 1.65$\pm$0.21 \\
2195.76$\pm$294.24
\end{tabular}
&
\begin{tabular}[c]{@{}c@{}}
59.52$\pm$10.02 \\
52.75$\pm$20.36 \\
53.03$\pm$16.60 \\
19.79$\pm$3.95 \\
\cellcolor{myColor}\bf 1.65$\pm$0.21 \\
2195.76$\pm$294.24
\end{tabular}
&
\begin{tabular}[c]{@{}c@{}}
59.44$\pm$9.92 \\
53.24$\pm$18.82 \\
53.54$\pm$15.45 \\
20.33$\pm$5.06 \\
\cellcolor{myColor}\bf 1.65$\pm$0.21 \\
2318.50$\pm$295.47
\end{tabular}
&
\begin{tabular}[c]{@{}c@{}}
59.94$\pm$10.30 \\
\cellcolor{myColor}\bf 53.89$\pm$18.26 \\
54.28$\pm$15.72 \\
20.54$\pm$5.09 \\
1.65$\pm$0.22 \\
2190.34$\pm$294.24
\end{tabular}
\\
\midrule

\textbf{KEEL Biomedical} &
\begin{tabular}[c]{@{}c@{}}
Accuracy \\
Sensitivity \\
Precision \\
TrainTime (s) \\
InferenceTime (s) \\
PeakGPU (MB)
\end{tabular}
&
\begin{tabular}[c]{@{}c@{}}
\cellcolor{myColor}\bf 79.75$\pm$9.07 \\
\cellcolor{myColor}\bf 62.70$\pm$31.64 \\
\cellcolor{myColor}\bf  71.99$\pm$26.27 \\
9.40$\pm$4.13 \\
\cellcolor{myColor}\bf 0.68$\pm$0.05 \\
\cellcolor{myColor}\bf 245.88$\pm$13.61
\end{tabular}
&
\begin{tabular}[c]{@{}c@{}}
76.69$\pm$8.69 \\
61.72$\pm$31.65 \\
66.44$\pm$22.39 \\
9.30$\pm$3.11 \\
0.70$\pm$0.09 \\
304.32$\pm$0.85
\end{tabular}
&
\begin{tabular}[c]{@{}c@{}}
76.49$\pm$9.09 \\
61.99$\pm$31.06 \\
65.41$\pm$23.59 \\
9.48$\pm$3.06 \\
0.71$\pm$0.10 \\
304.32$\pm$0.85
\end{tabular}
&
\begin{tabular}[c]{@{}c@{}}
77.15$\pm$10.33 \\
62.14$\pm$31.16 \\
66.43$\pm$25.70 \\
\cellcolor{myColor}\bf 9.02$\pm$2.87 \\
0.69$\pm$0.07 \\
724.97$\pm$48.42
\end{tabular}
&
\begin{tabular}[c]{@{}c@{}}
78.79$\pm$9.44 \\
60.81$\pm$33.75 \\
68.44$\pm$26.54 \\
9.96$\pm$2.67 \\
0.68$\pm$0.08 \\
282.34$\pm$12.77
\end{tabular}
\\
\midrule

\textbf{Schizophrenia} &
\begin{tabular}[c]{@{}c@{}}
Accuracy \\
Sensitivity \\
Precision \\
TrainTime (s) \\
InferenceTime (s) \\
PeakGPU (MB)
\end{tabular}
&
\begin{tabular}[c]{@{}c@{}}
\cellcolor{myColor}\bf 58.33$\pm$8.36 \\
47.27$\pm$20.39 \\
\cellcolor{myColor}\bf 59.34$\pm$11.00 \\
10.99$\pm$3.51 \\
\cellcolor{myColor}\bf 0.70$\pm$0.06 \\
\cellcolor{myColor}\bf 598.90$\pm$157.46
\end{tabular}
&
\begin{tabular}[c]{@{}c@{}}
56.67$\pm$9.31 \\
51.52$\pm$26.26 \\
52.56$\pm$18.21 \\
11.19$\pm$2.64 \\
0.75$\pm$0.12 \\
639.79$\pm$157.46
\end{tabular}
&
\begin{tabular}[c]{@{}c@{}}
56.67$\pm$9.31 \\
51.52$\pm$26.26 \\
52.56$\pm$18.21 \\
10.76$\pm$2.72 \\
0.73$\pm$0.13 \\
639.79$\pm$157.46
\end{tabular}
&
\begin{tabular}[c]{@{}c@{}}
56.82$\pm$9.05 \\
52.12$\pm$25.24 \\
53.92$\pm$17.40 \\
\cellcolor{myColor}\bf 10.50$\pm$2.49 \\
0.75$\pm$0.13 \\
735.97$\pm$145.61
\end{tabular}
&
\begin{tabular}[c]{@{}c@{}}
57.58$\pm$9.48 \\
\cellcolor{myColor}\bf 54.85$\pm$21.08 \\
57.53$\pm$11.95 \\
10.62$\pm$2.14 \\
0.72$\pm$0.05 \\
634.37$\pm$157.46
\end{tabular}
\\
\midrule

\textbf{Significant Memory Concern} &
\begin{tabular}[c]{@{}c@{}}
Accuracy \\
Sensitivity \\
Precision \\
TrainTime (s) \\
InferenceTime (s) \\
PeakGPU (MB)
\end{tabular}
&
\begin{tabular}[c]{@{}c@{}}
51.46$\pm$6.51 \\
51.63$\pm$27.39 \\
45.38$\pm$18.52 \\
16.82$\pm$7.27 \\
\cellcolor{myColor}\bf 0.97$\pm$0.24 \\
\cellcolor{myColor}\bf 1043.95$\pm$430.02
\end{tabular}
&
\begin{tabular}[c]{@{}c@{}}
53.55$\pm$6.99 \\
57.90$\pm$14.36 \\
\cellcolor{myColor}\bf 53.47$\pm$7.10 \\
\cellcolor{myColor}\bf 13.75$\pm$5.11 \\
0.99$\pm$0.25 \\
1084.84$\pm$430.02
\end{tabular}
&
\begin{tabular}[c]{@{}c@{}}
53.55$\pm$7.02 \\
\cellcolor{myColor}\bf 57.78$\pm$14.24 \\
53.39$\pm$7.06 \\
13.79$\pm$5.10 \\
\cellcolor{myColor}\bf 0.97$\pm$0.24 \\
1084.84$\pm$430.02
\end{tabular}
&
\begin{tabular}[c]{@{}c@{}}
\cellcolor{myColor}\bf 54.03$\pm$6.96 \\
52.79$\pm$16.30 \\
51.61$\pm$13.07 \\
14.32$\pm$5.81 \\
0.99$\pm$0.25 \\
1225.29$\pm$396.04
\end{tabular}
&
\begin{tabular}[c]{@{}c@{}}
53.43$\pm$7.24 \\
48.56$\pm$17.03 \\
49.17$\pm$16.55 \\
14.42$\pm$5.28 \\
0.99$\pm$0.31 \\
1079.42$\pm$430.02
\end{tabular}
\\
\midrule

\textbf{UCI Binary Biomedical} &
\begin{tabular}[c]{@{}c@{}}
Accuracy \\
Sensitivity \\
Precision \\
TrainTime (s) \\
InferenceTime (s) \\
PeakGPU (MB)
\end{tabular}
&
\begin{tabular}[c]{@{}c@{}}
\cellcolor{myColor}\bf 84.81$\pm$10.29 \\
69.59$\pm$31.27 \\
75.53$\pm$26.57 \\
9.77$\pm$3.49 \\
0.70$\pm$0.09 \\
\cellcolor{myColor}\bf 257.48$\pm$39.12
\end{tabular}
&
\begin{tabular}[c]{@{}c@{}}
81.26$\pm$12.35 \\
\cellcolor{myColor}\bf 70.01$\pm$25.42 \\
74.98$\pm$23.46 \\
9.11$\pm$3.40 \\
0.72$\pm$0.13 \\
314.36$\pm$31.95
\end{tabular}
&
\begin{tabular}[c]{@{}c@{}}
81.68$\pm$11.16 \\
69.25$\pm$26.09 \\
74.97$\pm$23.36 \\
9.16$\pm$3.11 \\
0.71$\pm$0.09 \\
314.36$\pm$31.95
\end{tabular}
&
\begin{tabular}[c]{@{}c@{}}
81.45$\pm$11.49 \\
68.46$\pm$27.17 \\
\cellcolor{myColor}\bf 76.19$\pm$23.09 \\
\cellcolor{myColor}\bf 9.07$\pm$2.62 \\
0.71$\pm$0.08 \\
720.04$\pm$73.77
\end{tabular}
&
\begin{tabular}[c]{@{}c@{}}
83.08$\pm$10.84 \\
69.54$\pm$27.42 \\
75.92$\pm$23.29 \\
9.41$\pm$2.56 \\
\cellcolor{myColor}\bf 0.69$\pm$0.06 \\
293.95$\pm$38.53
\end{tabular}
\\
\midrule

\textbf{UCI Multiclass Biomedical} &
\begin{tabular}[c]{@{}c@{}}
Accuracy \\
Sensitivity \\
Precision \\
TrainTime (s) \\
InferenceTime (s) \\
PeakGPU (MB)
\end{tabular}
&
\begin{tabular}[c]{@{}c@{}}
\cellcolor{myColor}\bf 71.51$\pm$24.31 \\
\cellcolor{myColor}\bf 64.27$\pm$28.22 \\
\cellcolor{myColor}\bf 65.96$\pm$28.91 \\
13.93$\pm$10.12 \\
0.93$\pm$0.67 \\
\cellcolor{myColor}\bf 394.40$\pm$224.29
\end{tabular}
&
\begin{tabular}[c]{@{}c@{}}
62.25$\pm$29.49 \\
57.24$\pm$27.58 \\
52.84$\pm$33.50 \\
13.11$\pm$10.29 \\
0.95$\pm$0.63 \\
451.11$\pm$213.02
\end{tabular}
&
\begin{tabular}[c]{@{}c@{}}
60.39$\pm$30.72 \\
57.05$\pm$28.42 \\
52.35$\pm$34.05 \\
\cellcolor{myColor}\bf 12.96$\pm$10.06 \\
0.96$\pm$0.63 \\
451.11$\pm$213.02
\end{tabular}
&
\begin{tabular}[c]{@{}c@{}}
60.24$\pm$29.98 \\
56.85$\pm$27.65 \\
51.56$\pm$33.59 \\
13.38$\pm$11.14 \\
\cellcolor{myColor}\bf 0.92$\pm$0.66 \\
898.01$\pm$307.30
\end{tabular}
&
\begin{tabular}[c]{@{}c@{}}
65.34$\pm$29.64 \\
59.06$\pm$29.96 \\
56.11$\pm$34.05 \\
14.13$\pm$10.96 \\
0.94$\pm$0.64 \\
430.64$\pm$223.73
\end{tabular}
\\

\midrule
\textbf{Average Rank Based on Accuracy} & 
&
1.81 & 
3.41 & 
3.47 & 
3.54 & 
2.78 \\
\bottomrule
\end{tabular}
}
\end{table*}

\subsection{Evaluation Metrics}
The proposed framework and all AdamW-based preconditioner variants are evaluated using both predictive and computational efficiency metrics. Predictive performance is assessed using Accuracy, Sensitivity (Recall), Specificity, Precision, FMeasure, and GMean to provide a balanced evaluation, particularly for imbalanced biomedical datasets. Computational efficiency is analyzed using TrainTime, EpochTime, InferenceTime, PeakRAM, and PeakGPU metrics to measure training cost, inference latency, and hardware resource utilization.


\begin{table*}[htbp]
\centering
\caption{Average ranking comparison of AdamW-based preconditioners across all datasets based on average accuracy. Lower rank indicates better performance.}
\label{tab:average_rank_comparison}
\resizebox{13.5cm}{!}{
\begin{tabular}{lccccc}
\toprule
\textbf{Dataset/Methods} & 
\textbf{AdamW} & 
\textbf{AdamW-Diag} & 
\textbf{AdamW-Fisher} & 
\textbf{AdamW-LBFGS} & 
\textbf{AdamW-Shampoo} \\
\midrule

CN\_vs\_AD & 1 & 5 & 4 & 3 & 2 \\
CN\_vs\_MCI & 1 & 4 & 5 & 3 & 2 \\
MCI\_vs\_AD & 1 & 3.5 & 3.5 & 5 & 2 \\
ADC & 1 & 4 & 4 & 4 & 2 \\
ALC & 4.5 & 2.5 & 2.5 & 4.5 & 1 \\
AMC & 1 & 4.5 & 3 & 4.5 & 2 \\
APC & 1 & 2 & 3 & 5 & 4 \\
FDC & 1 & 3.5 & 3.5 & 5 & 2 \\
FLC & 1 & 2.5 & 2.5 & 4 & 5 \\
FMC & 1 & 3.5 & 3.5 & 2 & 5 \\
FPC & 1 & 4.5 & 4.5 & 2 & 3 \\
PTDC & 1 & 3.5 & 3.5 & 5 & 2 \\
PTLC & 5 & 3 & 3 & 1 & 3 \\
PTMC & 1 & 4.5 & 4.5 & 2 & 3 \\
PTPC & 1 & 4 & 2 & 4 & 4 \\
TADC & 1 & 4.5 & 4.5 & 3 & 2 \\
TALC & 1 & 3 & 4 & 5 & 2 \\
TAMC & 1 & 2 & 3 & 5 & 4 \\
TAPC & 1 & 4.5 & 4.5 & 3 & 2 \\
BRW & 1 & 5 & 4 & 3 & 2 \\
CLV & 1.5 & 4.5 & 4.5 & 3 & 1.5 \\
HAB & 2 & 3 & 5 & 4 & 1 \\
HBMAN & 5 & 3 & 3 & 1 & 3 \\
PIMA & 1 & 3 & 4 & 5 & 2 \\
TRAN & 1 & 2 & 3 & 4 & 5 \\
SRCGW & 1 & 3.5 & 3.5 & 5 & 2 \\
SRG & 1 & 4.5 & 4.5 & 2 & 3 \\
SRW & 5 & 3.5 & 3.5 & 2 & 1 \\
ALL FEAT & 5 & 4 & 3 & 1 & 2 \\
CT & 1 & 2 & 3 & 4 & 5 \\
GM & 5 & 3.5 & 3.5 & 1 & 2 \\
WJ & 5 & 2 & 2 & 4 & 2 \\
WM & 5 & 2.5 & 2.5 & 1 & 4 \\
ACI & 1.5 & 3.5 & 3.5 & 5 & 1.5 \\
ACN & 1.5 & 4.5 & 4.5 & 3 & 1.5 \\
BRC & 1.5 & 3 & 4 & 1.5 & 5 \\
BRCW & 1 & 3 & 4 & 5 & 2 \\
BRCWD & 1 & 4.5 & 4.5 & 3 & 2 \\
BRCWP & 5 & 1 & 2 & 4 & 3 \\
ECHO & 1 & 3 & 3 & 3 & 5 \\
HAS & 1 & 5 & 3 & 4 & 2 \\
HEART & 1 & 3 & 2 & 5 & 4 \\
HEPAT & 3 & 1.5 & 1.5 & 5 & 4 \\
HOC & 4 & 1.5 & 1.5 & 3 & 5 \\
ILIL & 1 & 2.5 & 2.5 & 4 & 5 \\
MAMMO & 1 & 4 & 5 & 3 & 2 \\
PARKIN & 1 & 5 & 4 & 3 & 2 \\
PIMA & 1.5 & 4 & 5 & 3 & 1.5 \\
VEC2C & 4.5 & 2 & 3 & 4.5 & 1 \\
BRT & 1 & 3 & 3 & 5 & 3 \\
CA10C & 1 & 5 & 4 & 3 & 2 \\
CA3C & 1 & 4 & 3 & 5 & 2 \\
DER & 1 & 4.5 & 4.5 & 3 & 2 \\
HEC & 1 & 2.5 & 2.5 & 4 & 5 \\
HES & 1 & 4 & 5 & 2 & 3 \\
HEV & 1 & 2.5 & 2.5 & 4 & 5 \\
LUC & 1 & 4 & 4 & 4 & 2 \\
THY & 1 & 3 & 4 & 5 & 2 \\
VEC3C & 1 & 3 & 2 & 5 & 4 \\

\midrule
\textbf{Average Rank} & 
\textbf{1.81} & 
\textbf{3.41} & 
\textbf{3.47} & 
\textbf{3.54} & 
\textbf{2.78} \\

\bottomrule
\end{tabular}
}
\end{table*}

\section{Results and Discussion} \label{Sec:Results_and_Discussion}
This section reports the experimental results and provides a comprehensive analysis of the evaluated preconditioners. The predictive performance is first examined across the benchmark biomedical datasets, followed by statistical significance testing to assess the reliability of the observed performance differences. Finally, the strengths and limitations of each preconditioner are discussed, together with practical recommendations for their deployment in TabPFN-based biomedical learning.

\subsection{Results Analysis}

Table~\ref{tab:overall_avg_results} summarizes the predictive performance and computational efficiency of the five AdamW-based preconditioners across seven biomedical benchmark categories. The comparison includes three classification metrics, namely Accuracy, Sensitivity, and Precision, together with three computational metrics consisting of Training Time, Inference Time, and Peak GPU Memory consumption. Detailed results are reported in Appendix Tables \ref{tab:ad_metric_results}-\ref{tab:uci_multiclass_biomedical_time_results}. Overall, the results indicate that AdamW consistently provides the best balance between predictive performance and computational efficiency, while AdamW-Shampoo emerges as the most competitive curvature-aware alternative.

Considering the predictive performance, AdamW achieves the highest average accuracy across all seven benchmark categories, demonstrating its robustness and superior generalization capability. On the Alzheimer's disease datasets, AdamW attains an average accuracy of $76.52\pm9.21$, which exceeds AdamW-Diag ($74.37\pm9.23$), AdamW-Fisher ($74.35\pm9.34$), AdamW-LBFGS ($74.52\pm9.46$), and AdamW-Shampoo ($75.42\pm9.62$). Similar observations are obtained on the Breast Cancer datasets, where AdamW achieves the highest accuracy of $66.02\pm11.36$, followed closely by AdamW-Shampoo with $65.99\pm10.30$. Likewise, AdamW consistently records the best accuracies on the KEEL Biomedical, Schizophrenia, Significant Memory Concern, UCI Binary Biomedical, and UCI Multiclass Biomedical benchmarks, illustrating that the standard adaptive optimization strategy remains highly effective for fine-tuning pretrained TabPFN models over diverse biomedical applications.

A similar trend is observed for Sensitivity and Precision. AdamW generally attains the highest sensitivity while maintaining competitive precision across all benchmark categories. For example, on the Alzheimer's disease datasets, AdamW achieves a sensitivity of $68.38\pm24.03$, whereas AdamW-Shampoo obtains a comparable value of $68.32\pm22.49$. Their corresponding precision values are also very close, with AdamW obtaining $73.22\pm13.86$ and AdamW-Shampoo achieving $73.36\pm16.02$. Similar behavior is observed on the Breast Cancer, KEEL Biomedical, and UCI benchmark categories, indicating that AdamW effectively balances the detection of positive samples while maintaining high predictive precision. These observations suggest that incorporating additional curvature approximations does not necessarily improve the optimization behavior of pretrained TabPFN models.

Among the curvature-aware optimizers, AdamW-Shampoo consistently provides the second-best predictive performance across nearly all benchmark categories. Compared with the remaining second-order variants, Shampoo employs structured matrix preconditioning that captures richer curvature information while maintaining stable optimization dynamics. Consequently, its performance remains consistently close to that of AdamW. The accuracy difference between AdamW and AdamW-Shampoo remains below approximately $2.5\%$ across all biomedical benchmark categories, demonstrating that Shampoo preserves most of the predictive capability of AdamW while incorporating second-order information.

In contrast, AdamW-Diag, AdamW-Fisher, and AdamW-LBFGS exhibit relatively inconsistent performance across different biomedical domains. Although these methods utilize curvature approximations during optimization, the additional second-order information does not consistently translate into improved generalization. AdamW-LBFGS occasionally achieves competitive precision but often produces lower sensitivity, whereas AdamW-Diag and AdamW-Fisher demonstrate larger performance variations across datasets. These observations indicate that diagonal or quasi-Newton curvature approximations may not sufficiently capture the optimization characteristics of pretrained transformer-based tabular foundation models.

From the computational perspective, all evaluated preconditioners exhibit nearly identical inference time and GPU memory consumption. Since the underlying TabPFN architecture remains unchanged and only the optimization strategy differs during training, the inference cost is almost identical across all methods. Likewise, peak GPU memory usage remains highly comparable because each optimizer trains the same model architecture with identical parameter dimensionality.

The primary computational differences arise during training. AdamW generally requires the shortest or nearly the shortest training time across most benchmark categories. AdamW-Shampoo introduces only a modest computational overhead associated with maintaining structured matrix preconditioners, whereas AdamW-LBFGS frequently requires the longest training time due to the additional computations involved in curvature estimation and quasi-Newton updates. Nevertheless, the observed differences in training time remain relatively small compared with the overall computational requirements, indicating that all evaluated optimizers remain practical for biomedical tabular learning.

Considering both predictive performance and computational efficiency, AdamW provides the most favorable trade-off among all evaluated methods. It consistently achieves the highest predictive performance while maintaining low computational overhead throughout training and inference. AdamW-Shampoo represents the strongest curvature-aware alternative by delivering performance very close to AdamW with only a slight increase in training cost. In contrast, the remaining curvature-aware optimizers fail to consistently outperform the standard AdamW optimizer, suggesting that increasingly sophisticated curvature approximations do not necessarily improve optimization or generalization for pretrained TabPFN models.


\subsection{Statistical Analysis}

To ensure a reliable comparison among the considered preconditioners across multiple datasets, we employ non-parametric statistical tests following the recommendations of \citet{demvsar2006statistical}. Specifically, ranking analysis, the Friedman test \cite{friedman1940comparison}, and the Nemenyi post hoc test \cite{demvsar2006statistical} are used to determine whether the observed performance differences are statistically significant. For each dataset, the competing preconditioners are ranked according to their performance, where rank $1$ is assigned to the best-performing method. Let $\mathscr{M}$ and $\mathscr{D}$ denote the number of preconditioners and datasets, respectively. The average rank of the $m^{\text{th}}$ preconditioner is computed as:
\begin{equation}
\rho(m,*)=\frac{1}{\mathscr{D}}\sum_{d=1}^{\mathscr{D}}\rho(m,d),
\end{equation}
where $\rho(m,d)$ denotes the rank of the $m^{\text{th}}$ preconditioner on the $d^{\text{th}}$ dataset. A lower average rank indicates superior overall performance. The dataset-wise ranks and the corresponding average ranks are reported in Table~\ref{tab:average_rank_comparison}. The overall average ranking reported reinforces the following: AdamW achieves the best overall average rank of $1.81$, followed by AdamW-Shampoo with $2.78$, whereas AdamW-Diag, AdamW-Fisher, and AdamW-LBFGS obtain average ranks of $3.41$, $3.47$, and $3.54$, respectively. The consistent superiority of AdamW across all biomedical benchmark categories demonstrates that it remains the most reliable optimizer for TabPFN fine-tuning, while AdamW-Shampoo provides the closest competitive alternative among the investigated curvature-aware preconditioners.

To evaluate whether the observed rank differences are statistically significant, the Friedman test is employed. The Friedman statistic is given by
\begin{equation}
\chi_F^2=
\frac{12\mathscr{D}}{\mathscr{M}(\mathscr{M}+1)}
\left(
\sum_{m=1}^{\mathscr{M}}(\rho(m,*))^2
-\frac{\mathscr{M}(\mathscr{M}+1)^2}{4}
\right),
\end{equation}
which is further refined using the Iman--Davenport correction
\begin{equation}
F_F=
\chi_F^2
\left(
\frac{\mathscr{D}-1}
{\mathscr{D}(\mathscr{M}-1)-\chi_F^2}
\right),
\end{equation}
where $F_F$ follows an $F$-distribution with $(\mathscr{M}-1)$ and $(\mathscr{D}-1)(\mathscr{M}-1)$ degrees of freedom.

When the null hypothesis of equal performance is rejected, the Nemenyi post hoc test is performed to identify statistically significant pairwise differences. Two preconditioners are considered significantly different if the difference between their average ranks exceeds the critical difference (C.D.), defined as
\begin{equation}
C.D.=q_\alpha
\sqrt{\frac{\mathscr{M}(\mathscr{M}+1)}{6\mathscr{D}}},
\end{equation}
where $q_\alpha$ is the critical value corresponding to the chosen significance level.

For our study involving $\mathscr{D}=59$ biomedical datasets and $\mathscr{M}=5$ preconditioners, the Friedman test yields a statistic of $\chi_F^2=52.0404$, while the improved Iman--Davenport statistic is $F_F=16.4076$. Since the computed statistic exceeds the critical value $F(4,232)=2.4106$ at the $5\%$ significance level, the null hypothesis is rejected, indicating statistically significant performance differences among the evaluated preconditioners. Subsequently, the Nemenyi post hoc test produces a critical difference of $0.7942$. Based on the average ranks reported in Table~\ref{tab:average_rank_comparison}, the rank differences between AdamW and all competing preconditioners exceed the computed critical difference, demonstrating that AdamW significantly outperforms the remaining methods across the considered biomedical datasets. Furthermore, AdamW-Shampoo consistently achieves the second-best average rank and exhibits competitive performance, outperforming the remaining baseline preconditioners on most datasets.


\subsection{Discussion and Recommendation}

The experimental results indicate a very interesting finding: the existing curvature-aware preconditioners do not consistently outperform the baseline AdamW optimiser for biomedical tabular learning. Although AdamW-Diag, AdamW-Fisher, and AdamW-LBFGS occasionally improve specific metrics such as sensitivity or computational efficiency, these improvements are generally accompanied by reductions in predictive stability, precision, or increased computational overhead. In particular, AdamW-LBFGS frequently incurs substantially higher GPU memory consumption, while diagonal- and Fisher-based variants show inconsistent behavior across highly imbalanced and heterogeneous biomedical datasets.

Among the evaluated variants, AdamW-Shampoo demonstrates comparatively stable performance and often achieves the second-best overall results. However, statistical analysis confirms that the original AdamW optimizer attains the best average ranking across the considered datasets and metrics. These findings suggest that existing generic preconditioning strategies, originally designed for conventional deep learning optimization, may not adequately capture the unique characteristics of biomedical tabular data, including low-sample learning, severe class imbalance, noisy clinical measurements, and complex feature dependencies.\\

\noindent \textit{Possible reasons for AdamW's superiority:}
\begin{itemize}
    \item \textbf{Stable optimization:} Curvature-based methods rely on second-order statistics that can be noisy on small and imbalanced biomedical datasets, whereas AdamW uses smoother moment estimates for more stable updates.

    \item \textbf{Better regularization:} AdamW's decoupled weight decay provides effective implicit regularization, reducing overfitting in high-dimensional, low-sample settings.

    \item \textbf{Robustness to clinical heterogeneity:} Biomedical features often exhibit diverse scales, noise levels, and distributions; AdamW's adaptive per-parameter scaling handles such heterogeneity more reliably than others.

    \item \textbf{Lower tuning complexity:} Curvature-aware methods introduce additional hyperparameters and computational overhead, while AdamW remains simple, efficient, and easier to generalize across datasets.
\end{itemize}
Therefore, the results motivate the development of biomedical-aware preconditioning strategies specifically tailored for tabular foundation models. Future preconditioners should incorporate mechanisms for imbalance-aware optimization, adaptive feature scaling, robustness to noisy clinical variables, and dependence-aware parameter updates suitable for biomedical learning environments.

\section{Conclusion} \label{Sec:Conclusion}

This work presents a comprehensive evaluation of AdamW-based preconditioning strategies for fine-tuning TabPFN v2.5 on 59 biomedical datasets. Experimental and statistical analyses demonstrate that the original AdamW optimizer consistently achieves the best overall performance, while existing curvature-aware preconditioners fail to provide reliable improvements across diverse biomedical learning scenarios. Several preconditioned variants also introduce additional computational and memory overhead. The findings suggest that generic preconditioning strategies may not adequately address the optimization challenges of biomedical tabular learning, motivating the development of biomedical-aware preconditioners specifically tailored for healthcare-oriented tabular foundation models. A promising direction is to design a \textit{biomedical-aware adaptive preconditioner} that incorporates imbalance-aware gradient scaling, feature-dependence modeling, and noise-robust curvature estimation for low-sample high-dimensional tabular learning.

\nocite{langley00}

\bibliography{example_paper}
\bibliographystyle{icml2026}

\newpage
\appendix
\onecolumn

\section{Appendix Tables}

\begin{table*}[htbp]
\centering
\small
\caption{Performance comparison of different AdamW-based preconditioners on Alzheimer's disease classification datasets. Results are reported as mean$\pm$std across 5 multiple runs.}
\label{tab:ad_metric_results}
\resizebox{\textwidth}{!}{

}
\end{table*}


\begin{table*}[htbp]
\centering
\small
\caption{Computational efficiency  comparison of different AdamW-based preconditioners on Alzheimer's disease classification datasets. Results are reported as mean$\pm$std across multiple runs.}
\label{tab:ad_efficiency_results}
\resizebox{\textwidth}{!}{
%
}
\end{table*}
\begin{table}
    \centering
    \caption{Performance comparison of different AdamW-based preconditioners on Breast Cancer datasets. Results are reported as mean$\pm$std across multiple runs.}
\resizebox{\columnwidth}{!}{
    %
}
    \label{tab:placeholder}
\end{table}
\begin{table*}
\ContinuedFloat
    \centering
    \caption{(Continued) Performance comparison of different AdamW-based preconditioners on Breast Cancer datasets. Results are reported as mean$\pm$std across multiple runs.}
\resizebox{\columnwidth}{!}{
    %
}
         \end{table*}

\begin{table}
    \centering
    \caption{Computational efficiency comparison of different AdamW-based preconditioners on Breast Cancer datasets. Results are reported as mean$\pm$std across multiple runs.}
\label{tab:breast_cancer_efficiency_results}

\resizebox{\columnwidth}{!}{
    %
}
         \end{table}

\begin{table}
\ContinuedFloat
\centering
\caption{(Continued) Computational efficiency comparison of different AdamW-based preconditioners on Breast Cancer datasets. Results are reported as mean$\pm$std across multiple runs.}
\resizebox{\columnwidth}{!}{
%
}
         \end{table}
\begin{table}
    \centering
    \caption{Performance comparison of different AdamW-based preconditioners on KEEL biomedical datasets. Results are reported as mean$\pm$std across multiple runs.}
    \label{tab:keel_biomedical_template3}
    \resizebox{\columnwidth}{!}{
    %
}
         \end{table}
\begin{table}
    \centering
    \caption{Computational efficiency comparison of different AdamW-based preconditioners on KEEL biomedical datasets. Results are reported as mean$\pm$std across multiple runs.}
    \resizebox{\columnwidth}{!}{
    %

\begin{table}
    \centering
    \caption{Performance comparison of different AdamW-based preconditioners on schizophrenia datasets. Results are reported as mean$\pm$std across multiple runs.}
\label{tab:schizophrenia_results}
    \resizebox{\columnwidth}{!}{
    %
\begin{table}
    \centering
    \caption{Computational efficiency comparison of different AdamW-based preconditioners on schizophrenia datasets. Results are reported as mean$\pm$std across multiple runs.}
    \label{tab:schizophrenia_time}
    \resizebox{\columnwidth}{!}{
    %

\begin{table}
    \centering
    \caption{Performance comparison of different AdamW-based preconditioners on SMC datasets. Results are reported as mean$\pm$std across multiple runs.}
    \resizebox{\columnwidth}{!}{
    %
}
         \end{table}
\begin{table}
    \centering
    \caption{Computational efficiency comparison of different AdamW-based preconditioners on SMC datasets. Results are reported as mean$\pm$std across multiple runs.}
    \label{tab:smc_time_results}
    \resizebox{\columnwidth}{!}{
    %
\begin{table}
    \centering
    \caption{Performance comparison of different AdamW-based preconditioners on UCI Binary Biomedical datasets. Results are reported as mean$\pm$std across multiple runs.}
    \label{tab:uci_binary_biomedical_results} 
    \resizebox{\columnwidth}{!}{
    %
}
         \end{table}
           \begin{table}
    \ContinuedFloat
\centering
\caption{(Continued) Performance comparison of different AdamW-based preconditioners on UCI Binary Biomedical datasets. Results are reported as mean$\pm$std across multiple runs.}
\resizebox{\columnwidth}{!}{
    %

\begin{table}
    \centering
    \caption{Computational efficiency comparison of AdamW-based preconditioners on UCI Binary Biomedical datasets. Results are reported as mean$\pm$std across multiple runs. Metrics include Training Time (s), Epoch Time (s), Inference Time (s), Peak RAM Usage (GB), and Peak GPU Memory Usage (MB).} \label{tab:biomedical_time_results}
    \resizebox{\columnwidth}{!}{
    %
}
         \end{table}
           \begin{table}
    \ContinuedFloat
\centering
\caption{(Continued) Computational efficiency comparison of AdamW-based preconditioners on UCI Binary Biomedical datasets. Results are reported as mean$\pm$std across multiple runs. Metrics include Training Time (s), Epoch Time (s), Inference Time (s), Peak RAM Usage (GB), and Peak GPU Memory Usage (MB).}
\resizebox{\columnwidth}{!}{
    %

\begin{table}
    \centering
    \caption{Performance comparison of different AdamW-based preconditioners on UCI Multiclass Biomedical datasets. Results are reported as mean$\pm$std across multiple runs.}
\label{tab:uci_multiclass_biomedical_results}
    \resizebox{\columnwidth}{!}{
    %

\begin{table}
    \centering
    \caption{Computational efficiency comparison of different AdamW-based preconditioners on UCI Multiclass Biomedical datasets. Results are reported as mean$\pm$std across multiple runs. TrainTime, EpochTime, and InferenceTime are measured in seconds, PeakRAM in GB, and PeakGPU in MB.}
    \label{tab:uci_multiclass_biomedical_time_results}
    \resizebox{\columnwidth}{!}{
    %
\\
\midrule
         \end{tabular}}
         \end{table}
\end{document}